\documentclass[a4paper]{article}

\usepackage{tipa}
\usepackage[hidelinks]{hyperref}
\usepackage{INTERSPEECH2021}

\title{speechocean762: An Open-Source Non-native English Speech Corpus For Pronunciation Assessment}
\name{Junbo Zhang$^1$, Zhiwen Zhang$^2$, Yongqing Wang$^1$, Zhiyong Yan$^1$, \\
      Qiong Song$^2$, Yukai Huang$^2$, Ke Li$^2$, Daniel Povey$^1$, Yujun Wang$^1$}
\address{
  $^1$Xiaomi Corporation, Beijing, China\\
  $^2$SpeechOcean Ltd., Beijing, China}
\email{ \{zhangjunbo1, wangyongqing3, yanzhiyong, dpovey, wangyujun\}@xiaomi.com, \\
\{zhangzhiwen01, songqiong, huangyukai, like\}@speechocean.com}

\begin{document}

\maketitle
\begin{abstract}
  This paper introduces a new open-source speech corpus named “speechocean762” designed for pronunciation assessment use, consisting of 5000 English utterances from 250 non-native speakers, where half of the speakers are children. Five experts annotated each of the utterances at sentence-level, word-level and phoneme-level. A baseline system is released in open source to illustrate the phoneme-level pronunciation assessment workflow on this corpus. This corpus is allowed to be used freely for commercial and non-commercial purposes. It is available for free download from OpenSLR, and the corresponding baseline system is published in the Kaldi speech recognition toolkit.
\end{abstract}
\noindent\textbf{Index Terms}: corpus, computer-assisted language learning (CALL), second language (L2)

\section{Introduction}

As an indispensable part of Computer-aided language learning (CALL), computer-aided pronunciation training (CAPT) applications with pronunciation assessment technology are widely used in foreign language learning \cite{franco2010eduspeak,li2020training} and proficiency tests \cite{gu2020using}.
CAPT has been proved very useful to improve the pronunciation of the foreign language learners \cite{wang2014optimization}.
Due to the acute shortage of qualified teachers \cite{mcvey2019nuance} and the increasing popularity of online learning, the research of pronunciation assessment is being paid more attention \cite{cheng2020improving}.

According to the real-world CAPT applications' features, we divide the practical pronunciation assessment tasks into three categories by the assessment granularity: sentence-level, word-level, and phoneme-level.
The sentence-level assessment evaluates the whole sentence.
Specifically, three types of sentence-level scores frequently appear in practical CAPT systems: accuracy, completeness, and fluency.
The accuracy indicates the level of the learner pronounce each word in the utterance correctly; the completeness indicates the percentage of the words that are actually pronounced, and the fluency here is in the narrow sense\cite{lennon2000lexical}, which focuses on whether the speaker pronounces smoothly and without unnecessary pauses.
The word-level assessment has a finer scale than the sentence-level assessment. Typical word-level scores are accuracy and stress. Furthermore, as the finest granularity assessment, the phoneme-level assessment evaluates each phone's pronunciation quality in the utterance.
Note that the word-level accuracy score should not be regarded as the simple average of the phone-level accuracy scores, although they have strong correlations. Take the word ``above'' (/\textschwa \textprimstress b\textturnv v/) as an example. A foreign language learner may mispronounce it as /\textschwa \textprimstress b\textscripta v/ (mispronounce /\textturnv/ to /\textscripta/ ) or as /\textschwa \textprimstress k\textturnv v/ (mispronounce /b/ to /k/). For the two incorrect pronunciations, the numbers of the mispronounced phones are both one, but most people may realize that the latter mispronunciation is worse than the former.

There are some public corpora for pronunciation assessment. 
The ISLE Speech Corpus \cite{menzel2000isle} is an early and widely accepted \cite{oba2003using,honig2012automatic,papi2020mixtures} data set. It contains mispronunciation tags at the word and phoneme level, and the speakers are all from German and Italian. It is free for academic use, but it is charged for commercial use.
ERJ \cite{minematsu2004development} is another famous non-native English corpus for pronunciation assessment, collected from 202 Japanese students annotated with phonemic and prosodic symbols.
ATR-Gruhn \cite{gruhn2004multi} is a non-native English corpus with multiple accents. The annotations of ATR-Gruhn are speaker-level proficiency ratings.
TL-school \cite{gretter2020tlt} is a corpus of speech utterances collected in northern Italy schools for assessing the performance of students learning both English and German.
The data set of a spoken CALL shared task \cite{baur2017overview} is available to download, where Swiss students answer prompts in English, and the students' responses are manually labeled as ``accept'' or ``reject''.
L2-ARCTIC \cite{zhao2018l2} is a non-native English speech corpus with manual annotations, which has been used in some recent studies \cite{yan2020end,feng2020sed}, and it uses substitution, deletion, and insertion to annotate for the phoneme-level scoring.
Sell-corpus \cite{chen2019sell} is another multiple accented Chinese-English speech corpus with phoneme substitution annotations.
Some corpora, such as CU-CHLOE \cite{li2016mispronunciation}, Supra-CHLOE \cite{li2011design} and COLSEC \cite{yang2005construction}, have been used in many studies \cite{luo2011improvement,li2013lexical,li2017intonation,li2018automatic} but are not publicly available.
Corpora for languages other than English also exist. The Tokyo-Kikuko \cite{nishina2004development} is a non-native Japanese corpus with phonemic and prosodic annotations.
The iCALL corpus \cite{chen2015icall} is a Mandarin corpus spoken by non-native speakers of European descent with annotated pronunciation errors.
The SingaKids-Mandarin \cite{shang2012singapore} corpus focuses on mispronunciation patterns in Singapore children’s Mandarin speech.

To our knowledge, none of the existing non-native English corpora for pronunciation assessment contains all the following features:
\begin{itemize}
\item It is available for free download for both commercial and non-commercial purposes.
\item The speaker variety encompasses young children and adults.
\item The manual annotations are in many aspects at sentence-level, word-level and phoneme-level.
\end{itemize}

To meet these features, we created this corpus to support researchers in their pronunciation assessment studies.
The corpus is available on the OpenSLR \footnote{\url{https://www.openslr.org/101}} website, and the corresponding baseline system has been a part of the Kaldi speech recognition toolkit \footnote{\url{https://github.com/kaldi-asr/kaldi/tree/master/egs/gop\_speechocean762}}.

The rest of this paper is organized as follows: Section 2 describes the audio acquisition. Section 3 details how we annotated the data for the pronunciation assessment tasks. In Section 4, a Kaldi recipe for this corpus is introduced, which illustrates how to do phoneme-level pronunciation assessment, and the experiment results are provided as well.

\section{Audio Acquisition}
\label{sec:audio}

\begin{figure}[t]
  \centering
  \includegraphics[scale=0.2]{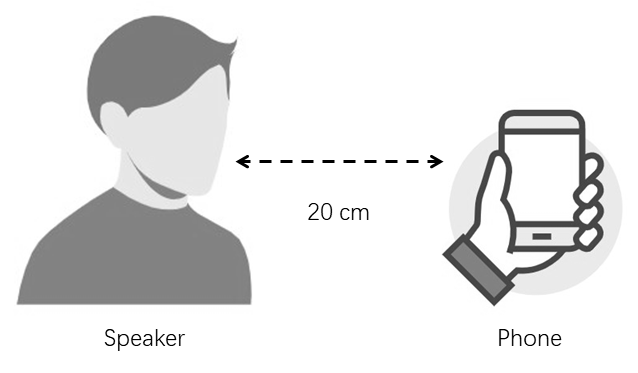}
  \caption{Recording setup. Speakers read the text holding their mobile phones in a quiet room.}
  \label{fig:recording}
\end{figure}

This corpus's text script is selected from daily life text, containing about 2,600 common English words. As shown in Figure \ref{fig:recording}, speakers were asked to hold their mobile phones 20cm from their mouths and read the text as accurately as possible in a quiet 3$\times$3 meters room. The mobile phones include the popular models of Apple, Samsung, Xiaomi, and Huawei. The number of sentences read aloud by each speaker is 20, and the total duration of the audio is about 6 hours.

The speakers are 250 English learners whose mother tongue is Mandarin.
The training set and test set are divided randomly, with 125 speakers for each.

\begin{figure}[t]
  \centering
  \includegraphics[scale=0.08]{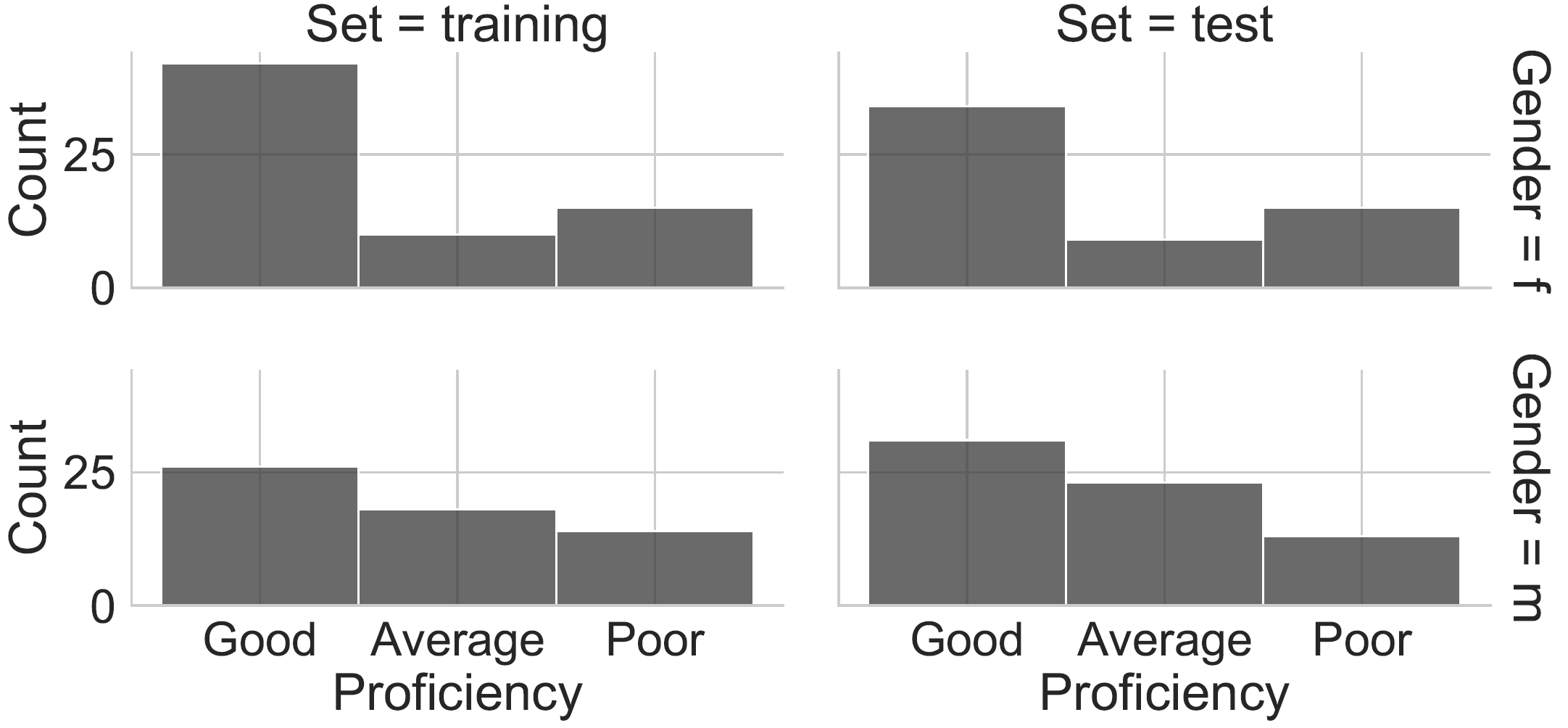}
  \caption{Speaker's English pronunciation proficiency distributions.}
  \label{fig:prof_dist}
\end{figure}

\begin{figure}[t]
  \centering
  \includegraphics[scale=0.135]{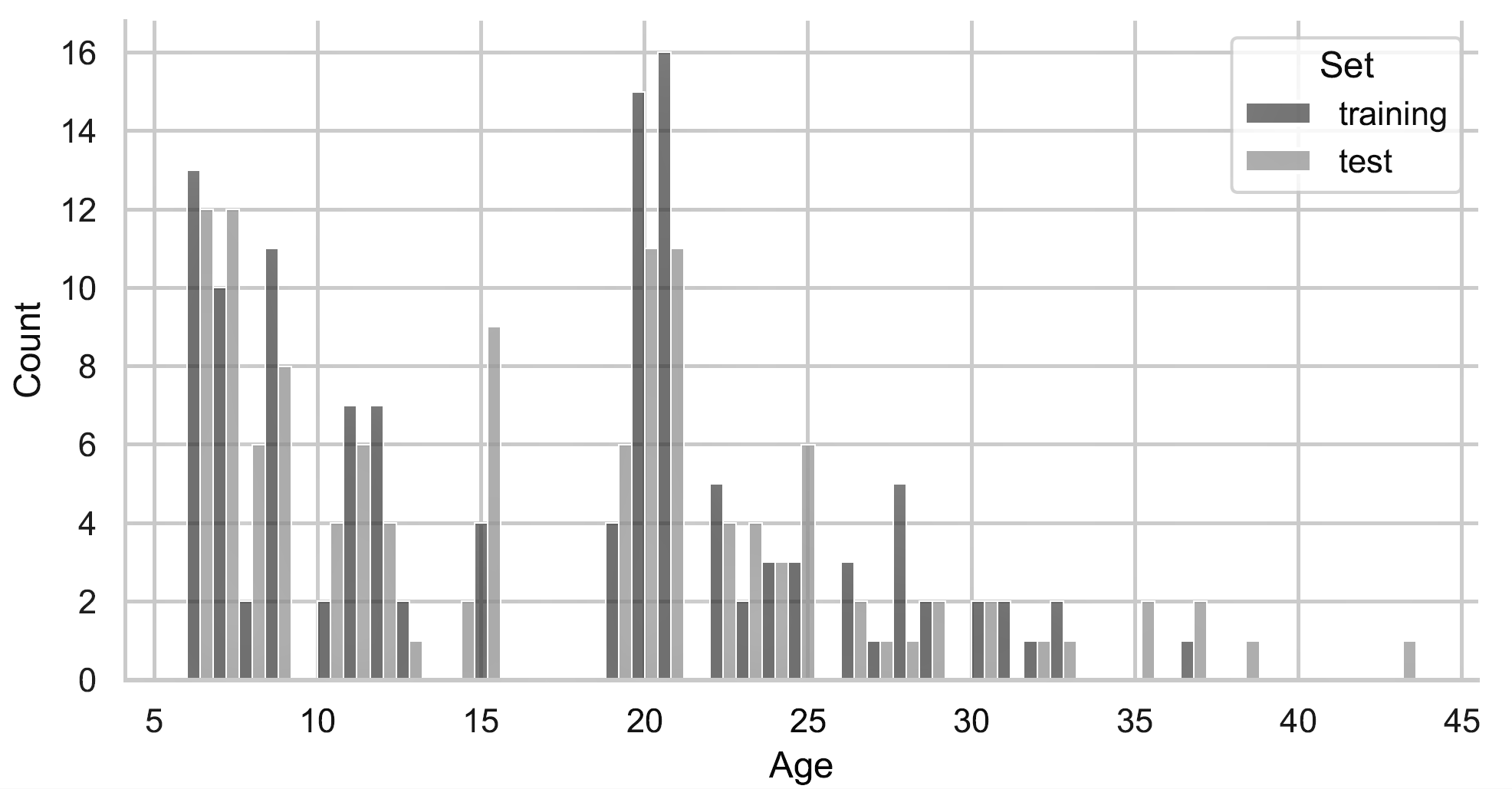}
  \caption{Speaker's age distributions.}
  \label{fig:age_dist}
\end{figure}

We carefully selected the speakers considering gender, age and proficiency of English.
The experts roughly rated the speaker's English pronunciation proficiency into three levels: good, average, and poor.
Figure \ref{fig:prof_dist} shows the distributions of the speaker's English pronunciation proficiency. 
Figure \ref{fig:age_dist} shows the distributions of the speaker's age. 
The gender ratio is 1:1 for both adults and children.

\begin{figure}[t]
  \centering
  \includegraphics[scale=0.25]{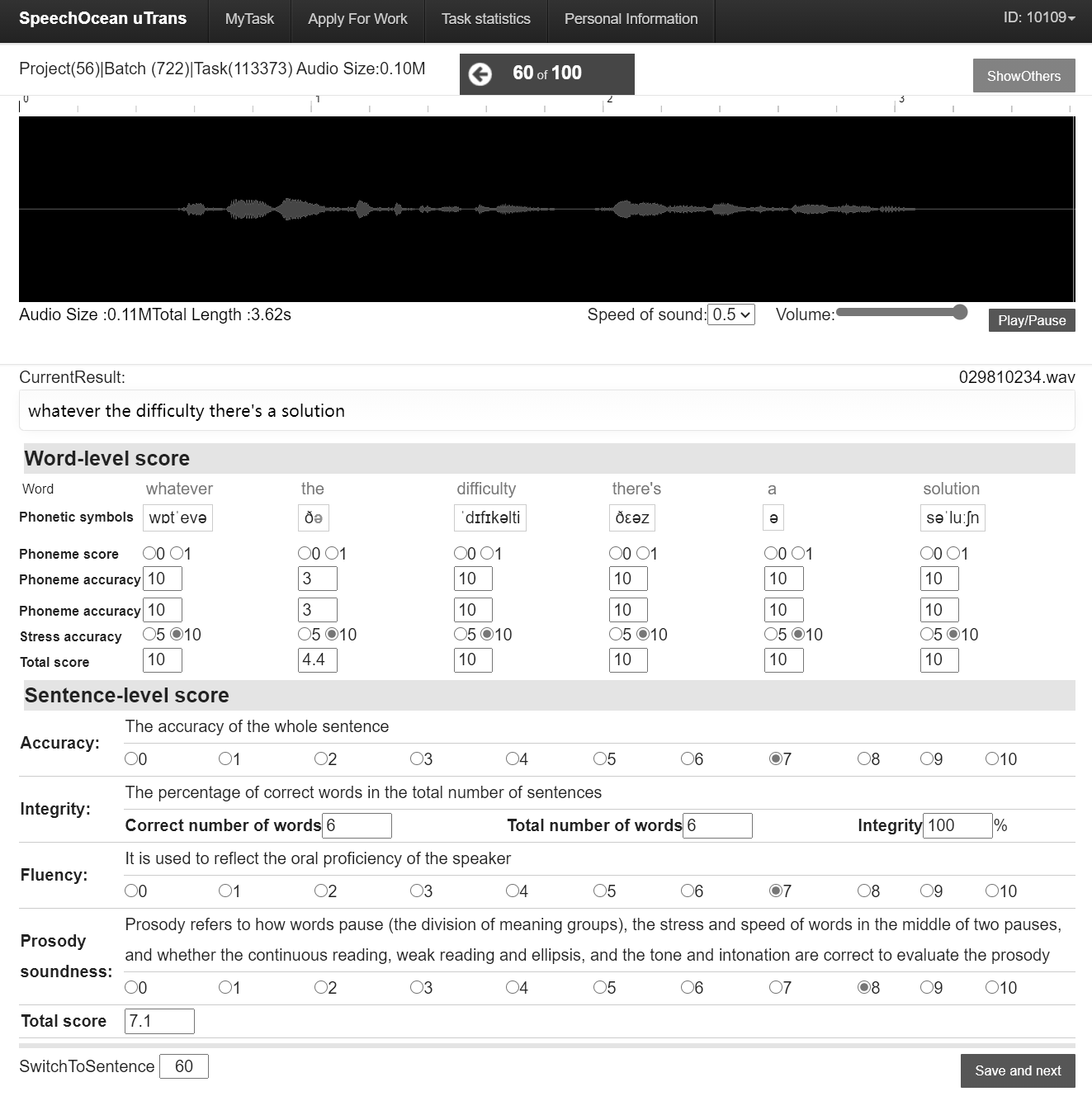}
  \caption{The ``SpeechOcean uTrans'' Application. Before this dialog is displayed, the experts have reached an agreement on the canonical phone sequences by voting. For the phoneme-level scoring, the expert selects the phone symbol and then makes a score of 0 or 1. If a phone symbol is not be selected, the score would be 2 as the default.}
  \label{fig:speech_production}
\end{figure}
\section{Manual Annotation}

Manual annotations are the essential part of this corpus. The annotations are the scores that indicate the pronunciation quality. Each utterance in this corpus is scored manually by five experts independently under the same metrics.

\subsection{Manual Scoring Metrics}
The experts discussed and formulated the manual scoring metrics.
Table \ref{tab:metrics} shows the detailed metrics.
The phoneme-level score is the pronunciation accuracy of each phone. The word-level scores include accuracy and stress, and the sentence-level scores include accuracy, completeness, fluency and prosody.
The sentence-level completeness score, which is not depicted in Table \ref{tab:metrics}, is the percentage of the words in the target text that are actually pronounced.

\begin{table*}[t]
  \caption{Manual Scoring Metrics}
  \label{tab:metrics}
  \centering
  \begin{tabular}{p{0.05\linewidth}p{0.8\linewidth}}
    \toprule
    \textbf{Score}   & \textbf{Description} \\
    \hline
    ~                & \textbf{Phoneme-level Accuracy} \\
    2                & The phone is pronounced correctly \\
    1                & The phone is pronounced with a heavy accent \\
    0                & The pronunciation is incorrect or missed \\
    \hline
    ~                & \textbf{Word-level Accuracy} \\
    10               & The pronunciation of the whole word is correct \\
    7-9              & Most phones in the word are pronounced correctly, but the word's pronunciation has heavy accents \\
    4-6              & No more than 30\% phones in the word are wrongly pronounced \\
    2-3              & More than 30\% phones in the word are wrongly pronounced, or be mispronounced into some other word \\
    0-1              & The whole pronunciation is hard to distinguish or the word is missed \\
    \hline
    ~                & \textbf{Word-level Stress} \\
    10               & The stress position is correct, or the word is a mono-syllable word \\
    5                & The stress position is incorrect \\
    \hline
    ~                & \textbf{Sentence-level Accuracy} \\
    9-10             & The overall pronunciation of the sentence is excellent without obvious mispronunciation \\
    7-8              & The overall pronunciation of the sentence is good, with a few mispronunciations \\
    5-6              & The pronunciation of the sentence has many mispronunciations but it is still understandable \\
    3-4              & Awkward pronunciation with many serious mispronunciations \\
    0-2              & The pronunciation of the whole sentence is unable to understand or there is no voice \\
    \hline
    ~                & \textbf{Sentence-level Fluency} \\
    8-10             & Coherent speech, without noticeable pauses, repetition or stammering \\
    6-7              & Coherent speech in general, with a few pauses, repetition and stammering \\
    4-5              & The speech is incoherent, with many pauses, repetition and stammering \\
    0-3              & The speaker is not able to read the sentence as a whole or there is no voice \\
    \hline
    ~                & \textbf{Sentence-level Prosodic} \\
    9-10             & Correct intonation, stable speaking speed and rhythm \\
    7-8              & Nearly correct intonation at a stable speaking speed \\
    3-6              & Unstable speech speed, or the intonation is inappropriate \\
    0-2              & The reading of the sentence is too stammering to do prosodic scoring or there is no voice \\
    \bottomrule
  \end{tabular}
\end{table*}

\begin{figure*}[t]
  \centering
  \includegraphics[width=\linewidth]{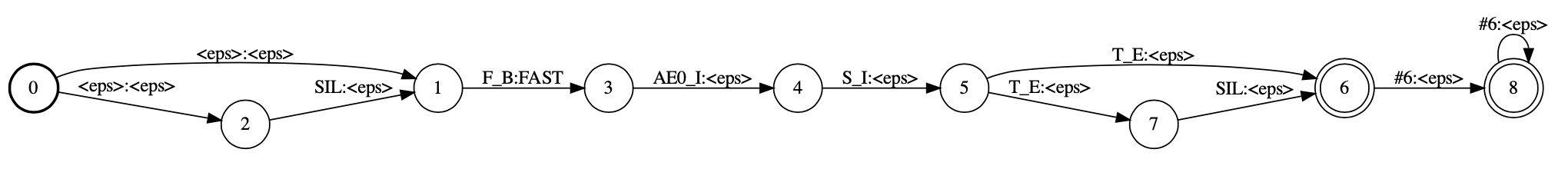}
  \caption{Building LG directly for the word ``fast'' with the canonical phone sequence voted by the experts, with skippable silence.}
  \label{fig:lg3}
\end{figure*}

\begin{figure}[t]
  \centering
  \includegraphics[scale=0.20]{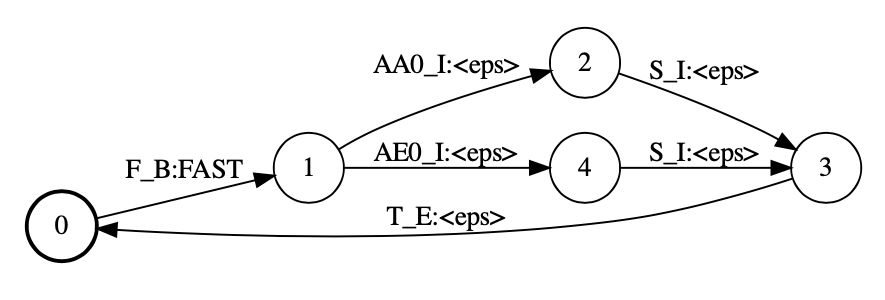}
  \caption{The part related of the word ``fast'' in L.}
  \label{fig:lg1}
\end{figure}

\subsection{The Multiple Canonical Phone Sequences Problem}
\label{sec:canonical}

The phoneme-level scoring requires determining the canonical phone sequence. A problem in practice is that the canonical phone sequence may not be unique. Take the word ``fast'' as an example. In middle school, most Chinese students were taught that this word should be pronounced as /f\textscripta :st/, so a proper canonical phone sequence is ``F AA S T'' with the phone set defined by the CMU Dictionary \cite{cmudict}. However, some speakers may pronounce this word as /f\ae st/ following the American pronunciation. If that is the case, the phone ``AA'' in the canonical phone sequence ``F AA S T'' would be misjudged as low score. The proper canonical phone sequence, in this case, should be ``F AE S T''.

Our solution is as follows. For each word, experts will be shown several possible canonical phone sequences before scoring. The expert must first select the sequence that is closest to the pronunciation in her or his belief. Since there are five experts, the sequence chosen by each expert may be different, so the five experts vote to determine the final canonical sequence. Then all the experts use the same canonical phone sequence to score. The canonical phone sequences are carried as a part of the corpus's meta-information.

\subsection{Scoring Workflow}

We developed an application named ``SpeechOcean uTrans'' for the experts to convieniently score the audio. The interface of the application is shown in Figure \ref{fig:speech_production}.

 Before the scoring, the experts read the transcript and listen to the audio to get familiar with the utterance. Then the experts are required to listen to the audio repeatedly at least three times. As we mentioned, some words have more than one canonical phone sequence. For those words, experts need to choose and vote to reach an agreement on the canonical phone sequence. Then the experts score the audio following the scoring metrics expressed in Table \ref{tab:metrics}. If the scores seem unreasonable, for example, the word-level score is high but all the phone-level scores are low, the ``SpeechOcean uTrans'' application would raise a warning message to remind the expert to recheck the scores.

\subsection{Score Distribution}
\label{sec:stat}
\begin{figure}[t]
  \centering
  \includegraphics[width=\linewidth]{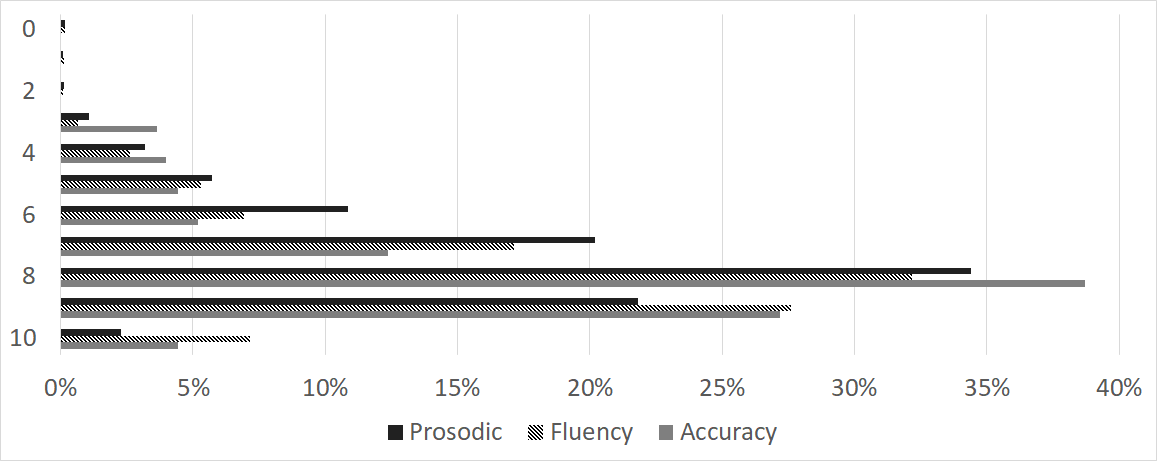}
  \caption{Sentence-level score distribution.}
  \label{fig:sen_scores}
\end{figure}

\begin{figure}[t]
  \centering
  \includegraphics[width=\linewidth]{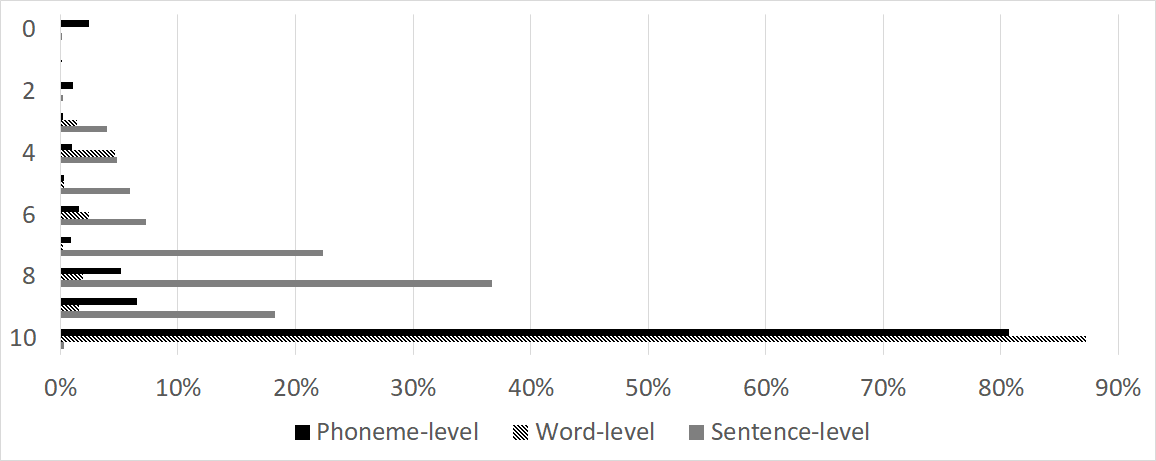}
  \caption{Score distribution in different levels.}
  \label{fig:multi_level_scores}
\end{figure}

Figure \ref{fig:sen_scores} shows the distribution of the sentence-level scores. 
The phoneme-level and word-level score distributions are shown in the Figure \ref{fig:multi_level_scores}, where the phoneme-level scores are mapped linearly to the range 0 to 10 for comparison.
The sentence-level scores variety encompasses 3 to 10, while most of the word-level and phoneme-level scores are from 8 to 10.
This behaviour stems from the fact that high sentence-level scores rely on a consistently ``good'' word and phoneme pronouncation.
Even a single word mispronunciation can lead to a low overall score.
Due to limited space, we suggest readers to refer to the available online corpus to obtain the detailed statistics.

\section{The Kaldi Recipe}

For demonstrating how to use this corpus to score at phoneme-level, we uploaded a recipe named ``gop\_speechocean762'' to the Kaldi toolkit.

\subsection{Pipeline}
We believe that the classical method is more suitable for building the baseline system than the latest methods.
So the pipeline is built following the neural network (NN) based goodness of pronunciation (GOP) method, which is widely used and detailed in \cite{hu2015improved}.
Here we only represent some specifics of implementing it on Kaldi.
The GOP method requires a pre-trained acoustic model trained by native spoken data, which is trained by the ``egs/librispeech/s5/local/nnet3/run\_tdnn.sh'' script in Kaldi.
The frame-level posterior matrix is generated through forward propagation on the native acoustic model, and the matrix is used for the forced alignment and the computing to obtain the GOP values and the GOP-based features, whose definitions could be found in \cite{hu2015improved} as well.
Then we train a regressor for each phone using the GOP-based features to predict the phoneme-level scores.

\subsection{Alignment Graph Building without Lexicon}

Kaldi's default alignment setup does not guarantee the alignment output to be identical to the canonical phone sequence voted by the experts. We continue to  use the word ``fast'' as the example. The two possible phone sequences of this word, which are ``F AA S T'' and ``F AE S T'' specifically, are both contained in the lexicon finite state transducer (FST), shown in Figure \ref{fig:lg1}. In that case, the phone sequence produced by the alignment is uncertain. If the experts' canonical phone sequence differs from the alignment result, the scores will not be comparable with the manual scores.

Therefore, we build the lexicon-to-grammar (LG) FST directly using the canonical phone sequence voted by the experts without composing the lexicon FST and the grammar FST. The process of directly constructing LG is simple: first, construct a linear FST structure, whose input labels are the canonical phone sequences voted by the experts, whereas the output labels are the corresponding words and epsilons \cite{mohri2008speech}. Then, add skippable silence between the words, and use the disambiguation symbol to construct the tail at the end of LG, as shown in Figure \ref{fig:lg3}.

\subsection{Supervised Training and Data Balancing}

With the GOP-based features and the corresponding manual scores, we train a regressor for each mono phone. The model structure is a support vector regressor (SVR) \cite{drucker1997support}.
Besides, we train polynomial regression models with the GOP values directly for each phone as an alternative lightweight method.

A problem is that the data's phoneme-level scores are quite unbalanced, as discussed in Section \ref{sec:stat}.
We use the high-score samples of other phones as the current phone's low-score samples to supplement the training set to address this issue. 
For example, a good pronunciation sample of the phone AE can be considered as a poor pronunciation sample of the phone AA. For the model training of a particular phone, we randomly select the samples of other phones with high manual scores, setting their scores as zero and add them to the training set.

\subsection{Results}

For evaluating the recipe's performance, we compare the predicted scores with the manual scores to calculate the mean squared error (MSE) and Pearson correlation coefficient (PCC). The result is shown in Table \ref{tab:result}.

As a baseline system, this recipe is based on the classical NN-based GOP method without using latest techniques.
So the result is not quite strong, which is in line with our expectations. 

\begin{table}[t]
  \caption{Performance of the recipe}
  \label{tab:result}
  \centering
  \begin{tabular}{lcc}
    \toprule
    \textbf{          }                          & \textbf{MSE}   & \textbf{PCC}  \\
    \midrule
    GOP value                                    &    0.69        &    0.25       \\
    GOP-based feature                            &    0.16        &    0.45       \\
    \bottomrule
  \end{tabular}
\end{table}

\section{Conclusions}
We released an open-source corpus for pronunciation assessment tasks. The corpus includes both child and adult speech and is manually annotated by five experts. The annotations are at sentence-level, word-level and phoneme-level. A Kaldi recipe is released to illustrate to use of the classic GOP method for phoneme-level scoring. In the future, we will expand the recipe to word-level and sentence-level scoring.

\section{Acknowledgements}
The authors would like to thank Jan Trmal for uploading this corpus to OpenSLR. The authors would also like to thank Heinrich Dinkel and Qinghua Wu for their helpful suggestions.


\bibliographystyle{IEEEtran}

\bibliography{mybib}

\begin{thebibliography}{10}
\providecommand{\url}[1]{#1}
\csname url@samestyle\endcsname
\providecommand{\newblock}{\relax}
\providecommand{\bibinfo}[2]{#2}
\providecommand{\BIBentrySTDinterwordspacing}{\spaceskip=0pt\relax}
\providecommand{\BIBentryALTinterwordstretchfactor}{4}
\providecommand{\BIBentryALTinterwordspacing}{\spaceskip=\fontdimen2\font plus
\BIBentryALTinterwordstretchfactor\fontdimen3\font minus
  \fontdimen4\font\relax}
\providecommand{\BIBforeignlanguage}[2]{{%
\expandafter\ifx\csname l@#1\endcsname\relax
\typeout{** WARNING: IEEEtran.bst: No hyphenation pattern has been}%
\typeout{** loaded for the language `#1'. Using the pattern for}%
\typeout{** the default language instead.}%
\else
\language=\csname l@#1\endcsname
\fi
#2}}
\providecommand{\BIBdecl}{\relax}
\BIBdecl

\bibitem{franco2010eduspeak}
H.~Franco, H.~Bratt, R.~Rossier, V.~Rao~Gadde, E.~Shriberg, V.~Abrash, and
  K.~Precoda, ``Eduspeak{\textregistered}: A speech recognition and
  pronunciation scoring toolkit for computer-aided language learning
  applications,'' \emph{Language Testing}, vol.~27, no.~3, pp. 401--418, 2010.

\bibitem{li2020training}
G.~Li, ``The training skills of college students’ oral {E}nglish based on the
  computer-aided language learning environment,'' in \emph{Journal of Physics:
  Conference Series}, vol. 1578, no.~1.\hskip 1em plus 0.5em minus 0.4em\relax
  IOP Publishing, 2020, p. 012040.

\bibitem{gu2020using}
L.~Gu, L.~Davis, J.~Tao, and K.~Zechner, ``Using spoken language technology for
  generating feedback to prepare for the {TOEFL iBT}{\textregistered} test: a
  user perception study,'' \emph{Assessment in Education: Principles, Policy \&
  Practice}, pp. 1--14, 2020.

\bibitem{wang2014optimization}
J.~Wang, ``On optimization of non-intelligence factors in college {E}nglish
  teaching in computer-aided language learning environments,'' in \emph{Applied
  Mechanics and Materials}, vol. 644.\hskip 1em plus 0.5em minus 0.4em\relax
  Trans Tech Publ, 2014, pp. 6124--6127.

\bibitem{mcvey2019nuance}
K.~P. McVey and J.~Trinidad, ``Nuance in the noise: The complex reality of
  teacher shortages.'' \emph{Bellwether Education Partners}, 2019.

\bibitem{cheng2020improving}
V.~C.-W. Cheng, V.~K.-T. Lau, R.~W.-K. Lam, T.-J. Zhan, and P.-K. Chan,
  ``Improving {E}nglish phoneme pronunciation with automatic speech recognition
  using voice chatbot,'' in \emph{International Conference on Technology in
  Education}.\hskip 1em plus 0.5em minus 0.4em\relax Springer, 2020, pp.
  88--99.

\bibitem{lennon2000lexical}
P.~Lennon, ``The lexical element in spoken second language fluency,'' in
  \emph{Perspectives on fluency}.\hskip 1em plus 0.5em minus 0.4em\relax
  University of Michigan, 2000, pp. 25--42.

\bibitem{menzel2000isle}
W.~Menzel, E.~Atwell, P.~Bonaventura, D.~Herron, P.~Howarth, R.~Morton, and
  C.~Souter, ``The {ISLE} corpus of non-native spoken {E}nglish,'' in
  \emph{Proceedings of LREC 2000: Language Resources and Evaluation Conference,
  vol. 2}.\hskip 1em plus 0.5em minus 0.4em\relax European Language Resources
  Association, 2000, pp. 957--964.

\bibitem{oba2003using}
T.~Oba and E.~Atwell, ``Using the {HTK} speech recogniser to anlayse prosody in
  a corpus of german spoken learner's {E}nglish,'' in \emph{UCREL Technical
  Paper number 16. Special issue. Proceedings of the Corpus Linguistics 2003
  conference}.\hskip 1em plus 0.5em minus 0.4em\relax Lancaster University,
  2003, pp. 591--598.

\bibitem{honig2012automatic}
F.~H{\"o}nig, T.~Bocklet, K.~Riedhammer, A.~Batliner, and E.~N{\"o}th, ``The
  automatic assessment of non-native prosody: Combining classical prosodic
  analysis with acoustic modelling,'' in \emph{Thirteenth Annual Conference of
  the International Speech Communication Association}, 2012.

\bibitem{papi2020mixtures}
S.~Papi, E.~Trentin, R.~Gretter, M.~Matassoni, and D.~Falavigna, ``Mixtures of
  deep neural experts for automated speech scoring,'' \emph{Proc. Interspeech
  2020}, pp. 3845--3849, 2020.

\bibitem{minematsu2004development}
N.~Minematsu, Y.~Tomiyama, K.~Yoshimoto, K.~Shimizu, S.~Nakagawa, M.~Dantsuji,
  and S.~Makino, ``Development of {E}nglish speech database read by {J}apanese
  to support call research,'' in \emph{Proceedings of ICA, vol. 1}.\hskip 1em
  plus 0.5em minus 0.4em\relax European Language Resources Association, 2004,
  pp. 557--560.

\bibitem{gruhn2004multi}
R.~Gruhn, T.~Cincarek, and S.~Nakamura, ``A multi-accent non-native {E}nglish
  database,'' in \emph{ASJ}, 2004, pp. 195--196.

\bibitem{gretter2020tlt}
R.~Gretter, M.~Matassoni, S.~Bann{\`o}, and F.~Daniele, ``{TLT}-school: a
  corpus of non native children speech,'' in \emph{Proceedings of The 12th
  Language Resources and Evaluation Conference}, 2020, pp. 378--385.

\bibitem{baur2017overview}
C.~Baur, C.~Chua, J.~Gerlach, E.~Rayner, M.~Russel, H.~Strik, and X.~Wei,
  ``Overview of the 2017 spoken call shared task,'' in \emph{Workshop on Speech
  and Language Technology in Education (SLaTE)}, 2017.

\bibitem{zhao2018l2}
G.~Zhao, S.~Sonsaat, A.~Silpachai, I.~Lucic, E.~Chukharev-Hudilainen, J.~Levis,
  and R.~Gutierrez-Osuna, ``{L2-ARCTIC}: A non-native {E}nglish speech
  corpus,'' \emph{Proc. Interspeech 2018}, pp. 2783--2787, 2018.

\bibitem{yan2020end}
B.-C. Yan, M.-C. Wu, H.-T. Hung, and B.~Chen, ``An end-to-end mispronunciation
  detection system for {L2} {E}nglish speech leveraging novel anti-phone
  modeling,'' in \emph{Proc. Interspeech 2020}, 2020, pp. 3032--3036.

\bibitem{feng2020sed}
Y.~Feng, G.~Fu, Q.~Chen, and K.~Chen, ``{SED-MDD}: Towards sentence dependent
  end-to-end mispronunciation detection and diagnosis,'' in \emph{IEEE
  International Conference on Acoustics, Speech and Signal Processing
  (ICASSP)}.\hskip 1em plus 0.5em minus 0.4em\relax IEEE, 2020, pp. 3492--3496.

\bibitem{chen2019sell}
Y.~Chen, J.~Hu, and X.~Zhang, ``Sell-corpus: an open source multiple accented
  chinese-english speech corpus for l2 english learning assessment,'' in
  \emph{IEEE International Conference on Acoustics, Speech and Signal
  Processing (ICASSP)}.\hskip 1em plus 0.5em minus 0.4em\relax IEEE, 2019, pp.
  7425--7429.

\bibitem{li2016mispronunciation}
K.~Li, X.~Qian, and H.~Meng, ``Mispronunciation detection and diagnosis in {L2}
  {E}nglish speech using multidistribution deep neural networks,''
  \emph{IEEE/ACM Transactions on Audio, Speech, and Language Processing},
  vol.~25, no.~1, pp. 193--207, 2016.

\bibitem{li2011design}
M.~Li, S.~Zhang, K.~Li, A.~M. Harrison, W.-K. Lo, and H.~Meng, ``Design and
  collection of an {L2} {E}nglish corpus with a suprasegmental focus for
  chinese learners of {E}nglish.'' in \emph{ICPhS}, 2011, pp. 1210--1213.

\bibitem{yang2005construction}
H.~Yang and N.~Wei, \emph{Construction and data analysis of a Chinese learner
  spoken {E}nglish corpus}.\hskip 1em plus 0.5em minus 0.4em\relax Shanhai
  Foreign Languse Eduacation Press, 2005.

\bibitem{luo2011improvement}
D.~Luo, X.~Yang, and L.~Wang, ``Improvement of segmental mispronunciation
  detection with prior knowledge extracted from large {L2} speech corpus,'' in
  \emph{Twelfth Annual Conference of the International Speech Communication
  Association}, 2011.

\bibitem{li2013lexical}
K.~Li, X.~Qian, S.~Kang, and H.~Meng, ``Lexical stress detection for {L2}
  {E}nglish speech using deep belief networks.'' in \emph{Interspeech}, 2013,
  pp. 1811--1815.

\bibitem{li2017intonation}
K.~Li, X.~Wu, and H.~Meng, ``Intonation classification for {L2} {E}nglish
  speech using multi-distribution deep neural networks,'' \emph{Computer Speech
  \& Language}, vol.~43, pp. 18--33, 2017.

\bibitem{li2018automatic}
K.~Li, S.~Mao, X.~Li, Z.~Wu, and H.~Meng, ``Automatic lexical stress and pitch
  accent detection for {L2} {E}nglish speech using multi-distribution deep
  neural networks,'' \emph{Speech Communication}, vol.~96, pp. 28--36, 2018.

\bibitem{nishina2004development}
K.~Nishina, Y.~Yoshimura, I.~Saita, Y.~Takai, K.~Maekawa, N.~Minematsu,
  S.~Nakagawa, S.~Makino, and M.~Dantsuji, ``Development of {J}apanese speech
  database read by non-native speakers for constructing call system,'' in
  \emph{Proc. ICA}, 2004, pp. 561--564.

\bibitem{chen2015icall}
N.~F. Chen, R.~Tong, D.~Wee, P.~Lee, B.~Ma, and H.~Li, ``i{CALL} corpus:
  Mandarin chinese spoken by non-native speakers of european descent,'' in
  \emph{Sixteenth Annual Conference of the International Speech Communication
  Association}, 2015.

\bibitem{shang2012singapore}
G.~Shang and S.~Zhao, ``Singapore mandarin: Its positioning, internal structure
  and corpus planning,'' in \emph{Paper presented atthe 22nd Annual Conference
  of the Southeast Asian Linguistics Society, Agay, France}, 2012.

\bibitem{cmudict}
R.~Weide, ``The {CMU} pronunciation dictionary.''\hskip 1em plus 0.5em minus
  0.4em\relax Carnegie Mellon University, 1998.

\bibitem{hu2015improved}
W.~Hu, Y.~Qian, F.~K. Soong, and Y.~Wang, ``Improved mispronunciation detection
  with deep neural network trained acoustic models and transfer learning based
  logistic regression classifiers,'' \emph{Speech Communication}, vol.~67, pp.
  154--166, 2015.

\bibitem{mohri2008speech}
M.~Mohri, F.~Pereira, and M.~Riley, ``Speech recognition with weighted
  finite-state transducers,'' in \emph{Springer handbook of speech
  processing}.\hskip 1em plus 0.5em minus 0.4em\relax Springer, 2008, pp.
  559--584.

\bibitem{drucker1997support}
H.~Drucker, C.~J. Burges, L.~Kaufman, A.~Smola, V.~Vapnik \emph{et~al.},
  ``Support vector regression machines,'' \emph{Advances in neural information
  processing systems}, vol.~9, pp. 155--161, 1997.

\end{thebibliography}

\end{document}